\documentclass[11pt,a4paper]{article}
\usepackage[table]{xcolor}

\usepackage{times,latexsym}
\usepackage{url}
\usepackage[T1]{fontenc}

\usepackage[acceptedWithA]{tacl2021v1}

\usepackage{xspace,mfirstuc,tabulary}

\usepackage[utf8]{inputenc}

\usepackage{microtype}

\usepackage{inconsolata}

\usepackage{graphicx}
\usepackage{booktabs}
\usepackage{multirow}
\usepackage{amsmath}
\usepackage{amssymb}
\usepackage{array}
\usepackage{enumitem}
\usepackage{cleveref}
\usepackage{CJKutf8}

\definecolor{lincolngreen}{rgb}{0.11, 0.35, 0.02}

\newcommand{\rankc}{\textsc{RankC}\xspace}

\newcommand{\cmark}{\checkmark}

\newcommand{\bxone}{{\boldsymbol{x}}_{1}}
\newcommand{\bxtwo}{{\boldsymbol{x}}_{2}}

\newcommand{\candone}{{\boldsymbol{c}}_{1}}
\newcommand{\candtwo}{{\boldsymbol{c}}_{2}}

\newcommand{\up}[1]{\textcolor{teal!70!black}{\scriptsize$\uparrow$#1}}
\newcommand{\down}[1]{\textcolor{red!75!black}{\scriptsize$\downarrow$#1}}

\newif\iftaclinstructions
\taclinstructionsfalse 
\iftaclinstructions
\renewcommand{\confidential}{}
\renewcommand{\anonsubtext}{(No author info supplied here, for consistency with
TACL-submission anonymization requirements)}
\newcommand{\instr}
\fi

\iftaclpubformat 

\else

\fi

\title{
A Systematic Evaluation of Cross-Lingual Consistency \\Enhancement Methods in Multilingual Language Models
}

\author{
  Jirui Qi$^{1}$ \
  Mingyang Wang$^{2}$ \
  Hinrich Schütze$^{2}$ \
  Raquel Fernández$^{3}$ \
  Arianna Bisazza$^{1}$ 
  \\
  $^1$University of Groningen \quad
  $^2$LMU Munich \quad
  $^3$University of Amsterdam 
  \\
  \texttt{j.qi@rug.nl} \quad
  \texttt{mingyang@cis.lmu.de} \quad
  \texttt{hinrich@hotmail.com} 
  \\
  \texttt{raquel.fernandez@uva.nl} \quad 
  \texttt{a.bisazza@rug.nl}
}

\date{}

\begin{document}
\maketitle
\begin{abstract}
Multilingual language models often produce inconsistent answers to semantically equivalent questions across languages, motivating methods to improve cross-lingual consistency (CLC). However, existing methods are typically evaluated using different models, tasks, and protocols, leaving their relative strengths unclear. In this work, we present a unified evaluation of representative CLC-enhancement methods for question answering, spanning inference-time interventions and post-training approaches across three model families and three closed-form benchmarks. The results show that post-training methods are generally more reliable, with direct distribution alignment consistently improving CLC across all model-dataset combinations, while other methods are more sensitive to answer format and the breadth of language coverage. Notably, cross-domain transfer is limited unless source and target tasks share similar output formats. We further investigate whether CLC enhancement hurts models' ability to respond differently \textit{when needed}, that is, when asked culture-dependent questions. Across two benchmarks of culturally diverse question answering, we find no systematic degradation in controlled closed-form evaluation, whereas open-ended generation reveals occasional accuracy reductions, particularly for non-English responses. Our work highlights the need to evaluate CLC enhancement for both cross-domain robustness and culturally appropriate variation, informing future work in post-training and benchmark development.
\footnote{All 
code and datasets will be released upon publication.}

\end{abstract}

\section{Introduction}
A reliable multilingual language model should provide consistent factual answers to semantically equivalent questions across languages. In practice, however, models often violate this expectation, producing substantially different answers across languages \citep{qi-etal-2023-cross, wang-etal-2025-lost-multilinguality, wang-etal-2025-calm}.
Such cross-lingual inconsistency can reduce answer accuracy, especially in low-resource languages, and undermine 
user trust.  

Recent studies have proposed diverse strategies for improving cross-lingual consistency (CLC). Some intervene directly in a model's internal representations at inference time \citep{wang-etal-2025-lost-multilinguality, wang-etal-2025-bridging, lu-etal-2025-paths}, while others post-train the model using English-centered supervision \citep{she-etal-2024-mapo, ICLR2025_6cdee724, zhang-etal-2025-cm}, multilingual preference signals \citep{wang-etal-2025-calm}, or explicit alignment of answer distributions across languages \citep{liu2026optimizing}. Despite promising results, these methods are typically evaluated with different base models, datasets, language sets, prompt formats, and protocols, making direct comparison difficult and leaving unclear which approaches are consistently effective. 

Beyond in-domain effectiveness, two practical questions remain underexplored. First, it is unclear whether CLC improvements learned on one task transfer to other tasks or knowledge domains. A method may align multilingual predictions within its post-training distribution without inducing broader multilingual knowledge consistency.
Understanding such generalization is important both for interpreting reported CLC gains and for determining how broadly alignment data must cover a target application.

Second, stronger consistency is not always desirable. While 
general factual questions have one single correct answer across languages, culturally or geographically grounded questions may legitimately require different answers. 
For instance, the question ``Should I leave a tip after dining at a restaurant?'' would typically be answered affirmatively in English but negatively in Japanese, reflecting different everyday norms associated with the two linguistic contexts.
An overly aggressive consistency objective may suppress such distinctions or shift non-English behavior toward an English-dominant answer distribution. To our knowledge, existing studies have not
directly examined this risk, particularly in controlled settings containing both culture-independent and culture-diverse knowledge.

In this work, we present a unified evaluation of representative CLC-enhancement methods for multilingual question answering. We cover three major paradigms: inference-time representation intervention, DPO-style post-training using English-centered or multilingual preference supervision, and direct alignment of answer distributions across languages. We evaluate them under shared question splits, prompts, candidate spaces, and metrics across three multilingual model families and five benchmarks.

Our results reveal three main findings. First, post-training methods are substantially more reliable than inference-time intervention, and direct distribution alignment provides the most stable gains across model--dataset setups.
Second, cross-domain transfer is generally weak and improves mainly when source and target tasks share similar answer formats, suggesting that current methods primarily align behavior near the post-training distribution rather than producing broad multilingual knowledge transfer. 
Third, we find that culturally diverse knowledge is largely preserved under CLC enhancement. In the closed-form evaluation, we observe no systematic degradation, and a follow-up probe shows that culture-independent and culture-diverse prompts are clearly separable in both lexical and hidden-state space. Overall accuracy also remains stable in open-ended generation, although we observe small drops for non-English responses, suggesting a potential risk of over-alignment in this setting.

Our contributions are as follows:
\begin{itemize}
\item We provide a controlled comparison of representative CLC-enhancement methods and characterize their relative strengths and limitations.

\item We systematically evaluate cross-domain transfer and show that CLC improvements generalize weakly unless source and target tasks share similar answer formats.

\item We present the first study of how CLC enhancement affects culture-diverse knowledge, finding no degradation in closed-form QA and clear representation separability between culture-independent and -diverse prompts. Accuracy in open-ended QA remains stable overall, despite small drops for non-English responses that suggest a risk of over-alignment.


\end{itemize}

\begin{table*}[!t]
\centering
\resizebox{\textwidth}{!}{
\begin{tabular}{lllccccc}
\toprule
\multirow{2}{*}{\textbf{Culture Category}}
& \multirow{2}{*}{\textbf{Dataset}}
& \multirow{2}{*}{\textbf{Task Category}}
& \multirow{2}{*}{\textbf{\# Langs.}}
& \multicolumn{2}{c}{\textbf{\# Samples}}
& \multirow{2}{*}{\textbf{QA Type}}
& \multirow{2}{*}{\textbf{Answer Format}} \\
\cmidrule(lr){5-6}
& & & & \textbf{Train} & \textbf{Test} & & \\
\midrule
\multirow{3}{*}{Independent} & BMLAMA & Factual knowledge & 17 & 5000 & 1792 & Closed & Text \\
& MMMLU & Multitask understanding & 14 & 5000 & 8942 & Closed & A/B/C/D \\
 & XCSQA & Commonsense reasoning & 16 & 800 & 200 & Closed & A/B/C/D/E \\
\midrule
\multirow{2}{*}{Independent \& diverse} & GEOMLAMA & Regional specific factual knowledge & 5 & - & 125 & Closed & Text \\
& BLEND & Local everyday knowledge & 7$^\dag$ & - & 500 & Open & Text\\
\bottomrule
\end{tabular}
}
\caption{Statistics of the evaluation datasets. Examples are provided in \Cref{app:dataset_examples}. $^\dag$ BLEND covers 13 languages in total, but we consider 7 in this work, as detailed in \Cref{sec:extension_blend}.
}
\label{tab:evaluation_datasets}
\end{table*}

\section{Related Work}
\label{sec:related_work}
\paragraph{Cross-lingual knowledge inconsistency.}
Prior work has shown that multilingual models often respond inconsistently to equivalent questions across languages \citep{kassner-etal-2021-multilingual, jiang-etal-2020-x, qi-etal-2023-cross}. In factual probing, the same relation--subject pair can yield different answer rankings depending on the prompt language. 
Metrics such as \rankc go beyond top-1 agreement by comparing the full ranking of candidate answers across languages \citep{qi-etal-2023-cross}, 
capturing both top-1 disagreement with similar overall rankings and top-1 agreement yet differing rankings over plausible alternatives.

\paragraph{Consistency enhancement methods.}
Several methods have been proposed to improve CLC, including inference-time representation intervention \citep{wang-etal-2025-lost-multilinguality, wang-etal-2025-bridging, lu-etal-2025-paths}, English-pivot preference alignment \citep{she-etal-2024-mapo, ICLR2025_6cdee724, zhang-etal-2025-cm}, multilingual self-alignment \citep{wang-etal-2025-calm}, direct consistency optimization \citep{liu2026optimizing}. Although these methods report promising gains, they are typically evaluated with different models, datasets, language sets, and protocols, making direct comparison difficult. We address this gap through a unified evaluation of representative methods.


\paragraph{Culturally diverse knowledge.}

Cross-lingual consistency is not always desirable: while culture-independent questions should generally yield the same answer across languages, culturally or geographically grounded questions may legitimately require different responses. This raises the risk that CLC enhancement could suppress culturally appropriate variation. 
Existing culturally grounded benchmarks, such as GEOMLAMA \citep{yin-etal-2022-geomlama} and BLEND \citep{myung2024blend}, provide a basis for evaluating such variation, yet it remains unclear whether CLC enhancement adversely affects culturally diverse knowledge. To our knowledge, this is the first work to systematically examine this underexplored question.

\section{Experimental Setup}
\label{sec:setups}

\subsection{Benchmarks and Framework}
\label{sec:benchmarks}

As discussed in \Cref{sec:related_work}, our evaluation covers both culture-independent and culturally diverse knowledge. We use five benchmarks with complementary roles, summarized in \Cref{tab:evaluation_datasets}.

\paragraph{Culture-independent benchmarks.}
Our \textbf{\textit{in-domain experiments}} (\Cref{sec:in-domain}) use three closed-form datasets whose questions are expected to share the same answer across languages. BMLAMA \citep{qi-etal-2023-cross} evaluates factual associations through cloze-style probing; MMMLU \citep{hendrycks2021measuring} evaluates multiple-choice knowledge across academic and professional subjects; and XCSQA \citep{lin-etal-2021-common} evaluates commonsense reasoning in a multiple-choice format. 
Despite covering different knowledge types, all three use finite candidate answer sets with shared gold answers across parallel prompts. We split each dataset at the level of parallel question sets to prevent semantically equivalent examples from appearing in both training and test sets. These datasets also support our \textbf{\textit{cross-domain experiments}} (\Cref{sec:cross_domain}), where MMMLU serves as the target for models post-trained on BMLAMA or XCSQA. 



\paragraph{Culturally diverse benchmarks.}
To test whether CLC enhancement harms culturally diverse knowledge (\Cref{sec:cultural_ood}), we use GEOMLAMA \citep{yin-etal-2022-geomlama}, whose geographically grounded prompts may require different answers across cultural contexts, making it a natural testbed for evaluating culturally diverse knowledge (\Cref{sec:geomalama-indomain}). BLEND \citep{myung2024blend} further complements this closed-form evaluation with open-ended, culturally grounded questions about everyday life (\Cref{sec:extension_blend}).
To better approximate realistic user--AI interactions, we exclude explicit country mentions from all prompts,
requiring models to infer the relevant locale from the query language and context.\footnote{See \Cref{app:dataset_examples} for examples.}

\begin{table*}[!t]
\centering
\resizebox{\textwidth}{!}{
\begin{tabular}{llllc}
\toprule
\textbf{Family} & \textbf{Methods} & \textbf{Update Type} & \textbf{Alignment Direction} & \textbf{Symmetric?} \\
\midrule
Representation intervention & INCLINE \citep{wang-etal-2025-bridging} & Inference-time & English $\rightarrow$ target & No \\
English-pivot preference alignment & MAPO / LIDR / CM-Align & Post-training & English $\rightarrow$ target & No \\
Multilingual self-alignment & CALM \citep{wang-etal-2025-calm} & Post-training & Multiway agreement & Yes \\
Direct consistency optimization & DCO \citep{liu2026optimizing} & Post-training & Pairwise or multiway & Yes \\
\bottomrule
\end{tabular}
}
\caption{
Representative methods for improving cross-lingual knowledge consistency. See \Cref{sec:method_family} for references to English-pivot preference alignment methods.}
\label{tab:method_families_new}
\end{table*}

\subsection{Evaluated CLC Method Families}
\label{sec:method_family}
We summarize representative CLC-enhancement methods in \Cref{tab:method_families_new}, grouped by training style and alignment directions. Note that our goal is not to propose a new training objective, but to compare existing enhancement strategies under a common protocol: the same data splits, prompt formats, candidate sets, metrics, and base models. 

\paragraph{Inference-time representation intervention.}

We use INCLINE \citep{wang-etal-2025-bridging} as the representative inference-time method because it is specifically designed for cross-lingual representation alignment. INCLINE learns layer-specific alignment matrices from parallel data to map lower-performing language representations toward a higher-performing language space, and applies these transformations to hidden states at inference time without updating model parameters.


\paragraph{English-pivot preference alignment (EN-Align).}
This family of methods relies on an asymmetric alignment signal that treats the English response as the preferred reference \citep{she-etal-2024-mapo, ICLR2025_6cdee724, zhang-etal-2025-cm}. In our experiments, we use MAPO as a representative method. When the English and target-language predictions disagree, the response selected in English is translated into the target language and used as the chosen response, while the original target-language response is used as the rejected response for DPO post-training.



\paragraph{Multilingual self-alignment.}
We adopt the CALM style self-alignment objective \citep{wang-etal-2025-calm} which derives preference signals from multilingual agreement rather than assuming English is correct. This objective favors responses supported across multiple languages and is naturally suited to joint-language post-training over parallel prompts.

\paragraph{Direct consistency optimization.}
DCO \citep{liu2026optimizing} directly optimizes CLC between ranked candidate distributions or response preferences. Given semantically equivalent prompts in different languages, it encourages the model to preserve the same answer preference ordering across languages: if a candidate is preferred under one language, it is trained to receive a similarly high preference under the others. Unlike CALM, which aligns toward a consensus answer from cross-lingual agreement, DCO imposes an explicit consistency objective and needs no majority vote or gold label. By targeting the full candidate ranking rather than a single preferred answer, it encourages consistency beyond top-1 predictions. \looseness=-1

\subsection{Training Setups}

We consider two training setups.\footnote{
Although intervention-based methods do not involve parameter training, they still require fitting weights on training splits. For convenience, we therefore refer to the corresponding stages as `training' when describing setups in this paper.} In \textbf{bilingual English-pivot post-training}, each training procedure contains English and one target language. This setup resembles practical scenarios where developers may be interested in aligning knowledge between English and a specific local language, or between a small set of regional languages. On the other hand, in \textbf{joint-language post-training}, each procedure contains all available language versions of the same question. This setup is closer to the scenario of building multilingual foundational models with more consistent cross-lingual knowledge.

To prevent data leakage from one language version of an item into another, we ensure that all language versions of the same question are assigned to the same split. Besides, all CLC-enhancement methods use the same training protocol, language sets, prompt templates, candidate answer sets, and base models for fair comparison.

\subsection{Evaluation Setups}
\label{sec:evalSetups}

\paragraph{Base Models}
We select advanced multilingual models from three families: Qwen2.5-7B-Instruct \citep{hui2024qwen2}, Gemma3-4B-IT \citep{Kamath2025Gemma3T}, Aya-Expanse-8B \citep{ustun-etal-2024-aya}, referred to as Qwen2.5, Gemma3, and Aya, respectively, in the rest of the paper. The base model is evaluated without any consistency post-training or intervention, providing the baseline for both accuracy and consistency.

\paragraph{Metrics} 

To measure \textit{accuracy} on the closed-form QA tasks (BMLAMA, MMMLU, XCSQA, and GEOMLAMA), we follow the LM-Evaluation-Harness protocol \citep{eval-harness} \footnote{\href{https://github.com/EleutherAI/lm-evaluation-harness}{github.com/EleutherAI/lm-evaluation-harness}}: candidate completions are ranked by model likelihood, and each prediction receives a score of 1 if the highest-ranked candidate matches the gold answer and 0 otherwise. Accuracy is then computed as the average score across all prompts. To measure
\textit{consistency},
we use the \rankc metric \citep{qi-etal-2023-cross}, which quantifies the consistency between candidate rankings induced by semantically equivalent prompts in different languages. We provide the formal definition of \rankc in \Cref{app:rankc}.

BLEND, our only open-ended benchmark, requires different metric formulations. For \textit{accuracy}, we follow the original paper and use their code for exact matching \citep{myung2024blend}. A response is considered correct if it matches any of the gold answers for the question. To account for morphological and orthographic variations, both annotations and model responses are normalized using language-specific lemmatization, stemming, or tokenization before matching. For
\textit{consistency},
\rankc cannot be adopted given the lack of candidate-answer lists. As a proxy, we therefore report the semantic similarity between the model responses across different languages, quantified using either BERTScore or cosine similarity.



\begin{table*}[!t]
\centering
\setlength{\tabcolsep}{1.6pt}
\renewcommand{\arraystretch}{1.08}
\resizebox{\textwidth}{!}{
\begin{tabular}{llcccccc|cccccc}
\toprule
\multirow{3}{*}{Model} & \multirow{3}{*}{Method}
& \multicolumn{6}{c|}{Bilingual English-pivot post-training}
& \multicolumn{6}{c}{Joint-language post-training} \\
\cmidrule(lr){3-8}\cmidrule(lr){9-14}
& & \multicolumn{3}{c}{CLC (\%)} & \multicolumn{3}{c|}{Accuracy (\%)}
& \multicolumn{3}{c}{CLC (\%)} & \multicolumn{3}{c}{Accuracy (\%)} \\
\cmidrule(lr){3-5}\cmidrule(lr){6-8}
\cmidrule(lr){9-11}\cmidrule(lr){12-14}
& & BMLAMA & MMMLU & XCSQA & BMLAMA & MMMLU & XCSQA
& BMLAMA & MMMLU & XCSQA & BMLAMA & MMMLU & XCSQA \\
\midrule
\multirow{5}{*}{Qwen2.5}
& Base
& 38.97 & 66.44 & 60.34 & 36.86 & 55.69 & 53.34
& 38.97 & 66.44 & 60.34 & 36.86 & 55.69 & 53.34 \\
& + INCLINE
& -3.23 & -0.11 & -0.32 & -2.77 & -0.04 & -0.39
& \multicolumn{6}{c}{\textit{Not applicable}} \\
& + CALM
& \multicolumn{6}{c|}{\textit{Not applicable}}
& +2.27 & +2.36 & +0.56 & +2.07 & +1.21 & +0.00 \\
& + EN-Align
& +17.73 & -7.92 & -7.23 & +13.44 & -11.32 & -12.64
& +20.99 & +1.18 & -1.38 & +18.14 & -0.33 & -13.38 \\
& + DCO
& +14.73 & +8.65 & +9.29 & +12.69 & +1.80 & +3.85
& +15.64 & +8.90 & +10.11 & +14.99 & +2.45 & +5.16 \\
\midrule
\multirow{5}{*}{Gemma3}
& Base
& 35.19 & 65.27 & 58.41 & 31.41 & 49.11 & 45.91
& 35.19 & 65.27 & 58.41 & 31.41 & 49.11 & 45.91 \\
& + INCLINE
& -0.72 & +0.06 & +0.08 & -0.41 & +0.05 & -0.29
& \multicolumn{6}{c}{\textit{Not applicable}} \\
& + CALM
& \multicolumn{6}{c|}{\textit{Not applicable}}
& +0.20 & +1.10 & +0.54 & +0.25 & -0.34 & -0.38 \\
& + EN-Align
& +1.31 & +5.33 & +0.77 & +1.06 & -3.80 & +0.41
& +2.14 & +5.32 & +0.88 & +1.84 & -2.44 & -1.47 \\
& + DCO
& +15.81 & +12.12 & +10.24 & +15.44 & -0.41 & +3.79
& +20.37 & +11.61 & +15.83 & +19.58 & +0.83 & +5.81 \\
\midrule
\multirow{5}{*}{Aya}
& Base
& 41.89 & 66.63 & 62.57 & 39.51 & 49.25 & 55.91
& 41.89 & 66.63 & 62.57 & 39.51 & 49.25 & 55.91 \\
& + INCLINE
& -15.67 & -0.04 & +0.01 & -15.99 & +0.00 & -0.35
& \multicolumn{6}{c}{\textit{Not applicable}} \\
& + CALM
& \multicolumn{6}{c|}{\textit{Not applicable}}
& +0.37 & +2.18 & +0.05 & +0.34 & -0.57 & -0.19 \\
& + EN-Align
& +4.94 & +3.46 & -1.25 & +4.54 & -2.95 & -3.61
& +11.57 & +0.22 & -0.09 & +9.80 & -0.61 & -1.09 \\
& + DCO
& +11.21 & +7.33 & +6.22 & +10.99 & +0.50 & +2.98
& +12.86 & +8.96 & +7.49 & +12.64 & +0.94 & +2.56 \\
\bottomrule
\end{tabular}
}
\caption{Evaluation results on BMLAMA, MMMLU, and XCSQA. We report both absolute scores and method-induced changes. Scores are averaged over all English--target language pairs. See \Cref{app:full_results} for detailed results.}
\label{tab:main_closed_form_summary}
\end{table*}

\section{Results and Findings}
\subsection{In-Domain Evaluation}
\label{sec:in-domain}

\Cref{tab:main_closed_form_summary} presents the main CLC and accuracy results on BMLAMA, MMMLU, and XCSQA. We organize the discussion around four main findings.



\paragraph{Finding 1: Post-training methods show more robust gains than inference-time representation intervention.}
Post-training methods generally yield more reliable CLC improvements across models and datasets. Particularly, DCO improves CLC in every model--dataset setting under both bilingual English-pivot and joint-language training, while CALM yields consistently positive, though smaller, gains under joint-language training. EN-Align is less consistent but still produces substantial improvements in several settings. In contrast, INCLINE, the inference-time intervention method in our comparison, does not yield robust gains.\footnote{INCLINE is applicable only in the bilingual setting since it operates through pairwise intervention over two languages.} Its effects are negligible on MMMLU and XCSQA and consistently negative on BMLAMA across all three models.
The degradation is especially pronounced for Aya, where where \rankc and accuracy drop by $15.67$ and $15.99$ points, respectively; Qwen2.5 also drops by $-3.23$ in \rankc and $-2.77$ in accuracy. 
These results diverge from the positive gains reported in the original INCLINE study. This could be due to its sensitivity to intervention configuration and alignment data, both of which have a substantial impact on performance \citep{wang-etal-2025-bridging}. This configuration dependence is consistent with the lower robustness we observe under our unified evaluation framework. Overall, post-training methods are more reliable than INCLINE.

\paragraph{Finding 2: DCO provides the most stable CLC improvements.}

Among all methods, DCO shows the most consistent gains, improving \rankc in every model--dataset combination under both bilingual English-pivot and joint-language post-training. Under joint-language training, \rankc increases by $+15.64$, $+8.90$, and $+10.11$ points for Qwen2.5 on BMLAMA, MMMLU, and XCSQA, respectively; by $+20.37$, $+11.61$, and $+15.83$ for Gemma3; and by $+12.86$, $+8.96$, and $+7.49$ for Aya. These consistency gains are generally accompanied by higher task accuracy, suggesting that DCO improves cross-lingual agreement while also shifting predictions toward more accurate answers, despite using no explicit gold-label supervision.


\begin{figure*}[!t]
    \centering
    \includegraphics[width=0.99\textwidth]{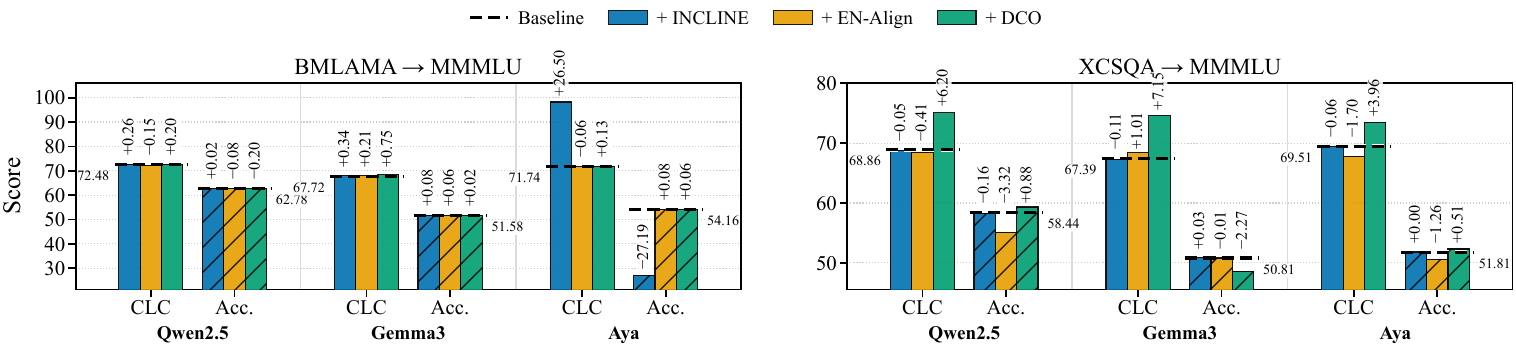}
    \caption{Average CLC score and accuracy of post-training on BMLAMA and XCSQA with evaluation on MMMLU. Solid bars indicate CLC whereas hatched bars indicate accuracy results. See \Cref{app:full_results} for full results.}
    \label{fig:cross_domain_summary}
\end{figure*}

\paragraph{Finding 3: English-pivot
preference alignment
can be effective but is format-sensitive.}

EN-Align
is more effective than INCLINE but less stable than DCO. It yields large gains on BMLAMA, especially for Qwen2.5, where joint-language training improves \rankc by $+20.99$ and accuracy by $+18.14$. However, performance is less consistent on the multiple-choice MMMLU and XCSQA. For Qwen2.5, bilingual English-pivot training reduces \rankc by $7.92$ on MMMLU and $7.23$ on XCSQA, with corresponding accuracy drops of $11.32$ and $12.64$. 
This pattern suggests that
asymmetric supervision anchored in English
is more effective when applied to semantically meaningful answer completions, such as concrete entities in factual probing, but can be brittle in multiple-choice settings where completions are represented by abstract option labels (e.g., A/B/C/D/E).



\paragraph{Finding 4: Multilingual self-alignment gives modest gains under noisy supervision.}


CALM-style multilingual self-alignment produces smaller but mostly positive consistency gains. Rather than treating English as the sole reference, CALM derives preferences from agreement across languages. However, its gains are more modest than those reported in the original CALM study \citep{wang-etal-2025-calm}.
This gap can be attributed to the reduced reliability of majority voting in a broader multilingual setting with more medium- and low-resource languages. As shown in \Cref{tab:calm_training_statistics}, the correctness ratio of CALM-selected completions is often only around 50\% across datasets and models. Such noisy preference pairs weaken DPO supervision by requiring the objective to fit potentially incorrect preferences while remaining close to the reference model, limiting improvements in both \rankc and answer accuracy. Taken together, these results underscore CALM's limitation in relying on reliable cross-lingual agreement, which may not hold in the practical and linguistically diverse setting.


\begin{table}[!h]
\centering
\small
\begin{tabular}{lccc}
\toprule
\textbf{Dataset} & \textbf{Qwen2.5} & \textbf{Gemma3} & \textbf{Aya} \\
\midrule
BMLAMA & 46.5\% & 38.6\% & 44.3\% \\
MMMLU & 52.6\% & 44.8\% & 47.9\% \\
XCSQA & 59.5\% & 51.2\% & 57.5\% \\
\bottomrule
\end{tabular}
\caption{Correctness ratio of the chosen completions in CALM training samples, averaged over all languages.}
\label{tab:calm_training_statistics}
\end{table}

\begin{figure*}[!t]
    \centering
    \includegraphics[width=0.99\textwidth]{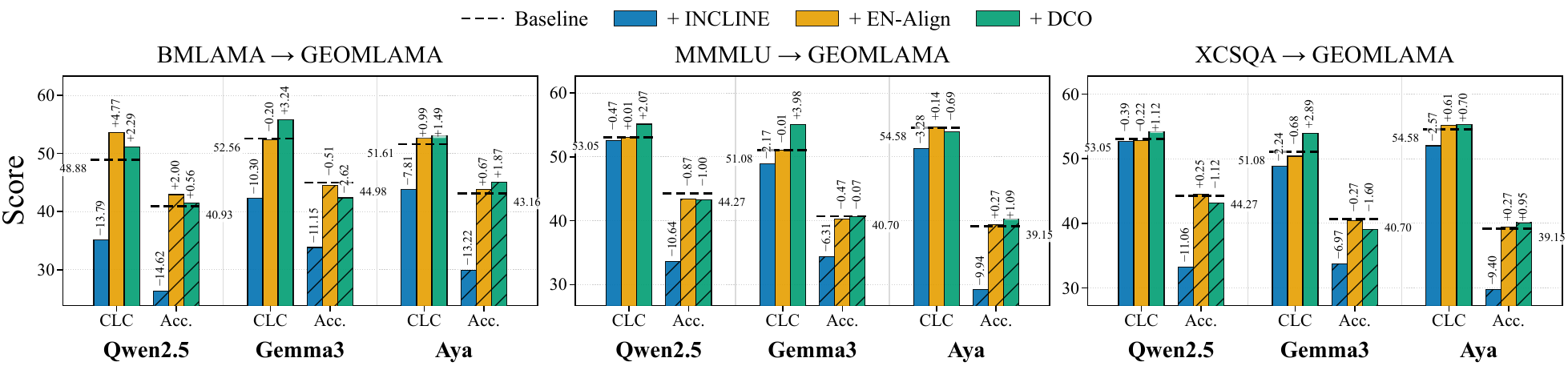}
    \caption{Evaluation on GEOMLAMA of models post-trained on culture-independent datasets. Solid bars indicate CLC whereas hatched bars indicate accuracy results.
    }
    \label{fig:geomlama_summary}
\end{figure*}

\subsection{Cross-Domain Generalization}

\label{sec:cross_domain}

The in-domain results show that CLC-enhancement methods can improve consistency across datasets and models. Next, we investigate whether these gains transfer beyond the post-training distribution, which is important for practical use on related downstream tasks.

We consider two cross-domain settings in which models are post-trained on either (1) BMLAMA or (2) XCSQA and then evaluated on MMMLU. Because the source and target datasets differ in language coverage, we adopt a bilingual training setup and restrict post-training and evaluation to English--target language pairs shared across the datasets.\footnote{We use \textsc{en} paired with {\textsc{fr}, \textsc{es}, \textsc{ar}, \textsc{ja}, \textsc{zh}, \textsc{ko}} for BMLAMA and {\textsc{zh}, \textsc{de}, \textsc{es}, \textsc{fr}, \textsc{it}, \textsc{ja}, \textsc{pt}, \textsc{ar}, \textsc{hi}, \textsc{sw}} for XCSQA. Accordingly, all subsequent experiments, including \Cref{sec:cultural_ood}, use the same bilingual setup, restricted to the languages overlapped between the corresponding training and test datasets.} This design controls for differences in language coverage and isolates cross-domain transfer.

As shown in \Cref{fig:cross_domain_summary}, BMLAMA$\rightarrow$MMMLU yields little useful transfer. 
EN-Align and DCO produce only minor changes in both \rankc and accuracy across all three models, in sharp contrast to their strong in-domain gains on BMLAMA in \Cref{tab:main_closed_form_summary}. This suggests that consistency improvements learned from factual probing do not readily generalize to broader multiple-choice knowledge evaluation.
INCLINE likewise fails to provide reliable transfer. Its effects are negligible for Qwen2.5 and Gemma3, while for Aya it increases \rankc by $+26.50$ but reduces accuracy by $-27.19$. This reflects degenerate consistency: rather than improving knowledge retrieval, the model collapses toward nearly fixed option-letter predictions rather than retrieving knowledge more reliably, causing accuracy to fall to about $25\%$, close to chance level on four-choice MMMLU questions.


The XCSQA$\rightarrow$MMMLU setting shows a different pattern. EN-Align remains unstable, with mixed CLC changes and accuracy drops for Qwen2.5 and Aya. In contrast, DCO improves \rankc for all three models, by $+6.20$ for Qwen2.5, $+7.15$ for Gemma3, and $+3.96$ for Aya. Although smaller than the corresponding in-domain gains, these improvements suggest that some CLC enhancement transfers more readily when the source and target datasets share a similar answer format. In particular, both XCSQA and MMMLU are multiple-choice tasks whose completions are option letters.\footnote{MMMLU uses A/B/C/D, while XCSQA uses A/B/C/D/E.}
However, even in this more favorable setting, transfer remains substantially weaker than in-domain. DCO therefore appears to transfer some format-level ranking alignment from XCSQA to MMMLU, but not the full consistency gains learned during post-training. More broadly, CLC enhancement learned on one dataset does not automatically generalize across knowledge domains, even when the response formats are similar.

These findings put the large in-domain gains of \Cref{tab:main_closed_form_summary} into perspective. Such improvements should not be interpreted as evidence of a global enhancement of multilingual knowledge consistency; rather, they primarily reflect better alignment within the post-training task distribution. Cross-domain generalization remains a separate challenge, highlighting the importance of matching alignment data to the intended use case and, more broadly, of using diverse training coverage when robust transfer is required.

\section{
Impact of CLC Enhancement on Culturally Diverse Knowledge
}
\label{sec:cultural_ood}
Beyond cross-domain generalization, CLC enhancement methods raise a second important concern: they may inadvertently over-align multilingual behavior where variation is desirable. By encouraging semantically similar responses across languages, these methods may suppress culturally specific knowledge whose appropriate expression depends on cultural or geographic context.

\subsection{Evaluation on GEOMLAMA}
\label{sec:geomalama-indomain}

\paragraph{Full dataset evaluation.} We first examine this risk on GEOMLAMA \citep{yin-etal-2022-geomlama}, which contains geographically and culturally grounded factual prompts in a format similar to BMLAMA. 
We evaluate models post-trained on the culture-independent BMLAMA, MMMLU, and XCSQA datasets directly on GEOMLAMA. Consistency gains paired with a substantial accuracy drop would indicate that CLC enhancement impairs the preservation of culturally grounded knowledge, whereas stable accuracy would suggest that improving CLC on culture-independent data does not necessarily collapse culturally diverse facts into a single language-independent answer distribution.

For the post-training methods, changes in GEOMLAMA CLC are generally modest. As shown in \Cref{fig:geomlama_summary}, DCO improves \rankc in most settings, but much less so than in-domain, ranging from $+0.70$ to $+3.98$, except for Aya after MMMLU post-training, where \rankc decreases by $-0.69$. EN-Align is even closer to neutral, with the exception of Qwen2.5 post-trained on BMLAMA, where \rankc improves by $+4.77$. This aligns with our finding in \Cref{sec:cross_domain} that CLC gains transfer more effectively when the training and test sets share similar response formats.

More importantly, these consistency changes are not accompanied by systematic degradation in accuracy. EN-Align yields only small accuracy changes across source datasets and models ($-0.87$ to $+2.00$), while DCO shows similarly mixed effects, including several improvements and a few moderate drops (e.g., $-2.62$ for Gemma3 after BMLAMA post-training). Overall, neither method consistently reduces accuracy on GEOMLAMA, suggesting that improving CLC through post-training in culture-independent domains does not necessarily harm culturally diverse factual knowledge.



The inference-time intervention INCLINE behaves differently. It consistently reduces GEOMLAMA accuracy across all source datasets and models, with especially large drops under BMLAMA-based intervention. After BMLAMA training, INCLINE reduces accuracy by $-14.62$, $-11.15$, and $-13.22$ for Qwen2.5, Gemma3, and Aya, respectively. Given INCLINE's instability in both the in-domain and cross-domain experiments, these drops are better understood as a consequence of brittle intervention than of over-alignment. Directly perturbing hidden-states may disrupt the model's normal prediction process.

\paragraph{Evaluation on the culturally diverse subset.}
To directly assess whether CLC enhancement harms culturally diverse knowledge, we conduct a controlled evaluation on GEOMLAMA by selecting English--target language pairs whose gold answers differ across languages. 
We focus on Qwen2.5 post-trained on BMLAMA, which yields the strongest CLC gains in \Cref{fig:geomlama_summary}.

\begin{table}[!t]
\centering
\setlength{\tabcolsep}{3.2pt}
\resizebox{\columnwidth}{!}{
\begin{tabular}{lrrr|rrr}
\toprule
\multirow{2}{*}{Method} &
\multicolumn{3}{c}{en--zh} &
\multicolumn{3}{c}{en--fa} \\
\cmidrule(lr){2-4}\cmidrule(lr){5-7}
& CLC & $\rm Acc_{en}$ & $\rm Acc_{zh}$
& CLC & $\rm Acc_{en}$ & $\rm Acc_{fa}$ \\
\midrule
Baseline
& 61.37 & 36.00 & 56.00
& 25.57 & 21.67 & 15.83 \\
EN-Align
& +3.62 & +0.00 & -1.50
& +1.83 & +0.00 & +4.17 \\
DCO
& +4.80 & -0.17 & +0.50
& +2.38 & +0.55 & +1.67 \\
\bottomrule
\end{tabular}
}
\caption{
CLC (\%) and accuracy (\%) on GEOMLAMA for Qwen2.5 post-trained on BMLAMA.
}
\label{tab:geomlama_only_diverse}
\end{table}

\Cref{tab:geomlama_only_diverse} shows that these gains transfer from BMLAMA post-training to GEOMLAMA without systematically reducing accuracy on culturally diverse questions. Accuracy changes are generally small, with the largest decrease being only $1.50$ from a baseline of $56.00$. For the English--Persian pair, EN-Align even improves Persian accuracy by $4.17$.

Taken together, these results suggest that a trade-off between CLC and culturally diverse knowledge is not inevitable: post-training can improve consistency on culture-independent data without systematically degrading GEOMLAMA accuracy. 
One possible explanation is that culture-diverse queries differ from culture-independent ones in their lexical content and internal representations, causing post-training updates to transfer differently. We examine this hypothesis further in \Cref{sec:separability_analysis}.

\subsection{Separability of Culture-Independent and Culture-Diverse Queries}
\label{sec:separability_analysis}

If models already encode culture-independent and culture-diverse questions distinctly before CLC enhancement, updates learned from culture-independent examples may transfer only weakly to culture-diverse questions.

To examine this hypothesis, we train linear probes to distinguish culture-independent from culture-diverse questions. We represent each question using two types of information: (i) lexical information computed with TF--IDF \citep{salton1988term} and Word2Vec \citep{iyyer-etal-2015-deep} and (ii) the information encoded in the representations by the LLMs we evaluate, extracted from each layer's final-token hidden states.\footnote{We follow prior work that treats final-token states as compact representations of the input \citep{liu-etal-2024-aligning, li-etal-2025-adaptive, bricken2023monosemanticity, levinstein2025still}.} For each question, we combine the English and the target-language representations using different strategies, including concatenation, mean pooling, absolute difference, and mean-plus-difference.


\begin{figure*}[!t]
\centering
\includegraphics[width=0.99\linewidth]{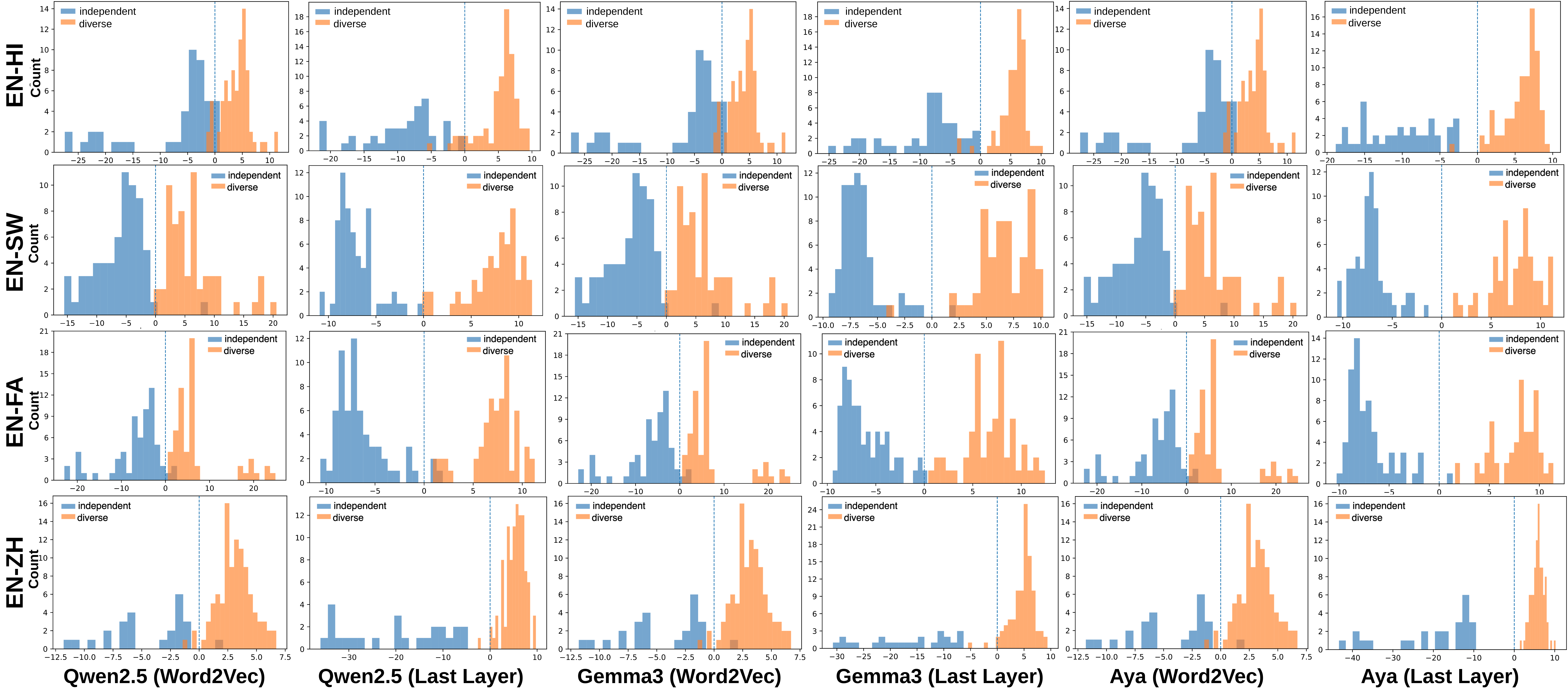}
\caption{
Linear-probe decision scores from lexical representations and hidden states for English--non-English query pairs. The probe decision scores before sigmoid transformation are presented for clearer visualization.
}
\label{fig:probe_scores}
\end{figure*}

\begin{table}[!t]
\centering
\resizebox{\columnwidth}{!}{
\begin{tabular}{lcccc}
\toprule
Subset & en--zh & en--hi & en--fa & en--sw \\
\midrule
Culture-indep. & 25 & 45 & 65 & 65 \\
Culture-diver. & 100 & 80 & 60 & 60 \\
\bottomrule
\end{tabular}
}
\caption{Sizes of the culture-independent and culture-diverse subsets for each English--target language pair in GEOMLAMA.}
\label{tab:geomlama_category_coverage}
\end{table}

\begin{table*}[!t]
\centering
\small

\begin{tabular}{lcccccc}
\toprule
\multirow{2}{*}{Post-training Data} &
\multicolumn{2}{c}{Qwen2.5} &
\multicolumn{2}{c}{Gemma3} &
\multicolumn{2}{c}{Aya} \\
\cmidrule(lr){2-3}
\cmidrule(lr){4-5}
\cmidrule(lr){6-7}
& BERTScore & CosSim
& BERTScore & CosSim
& BERTScore & CosSim \\
\midrule

BMLAMA
& 85.47 \up{0.25}
& 60.72 \up{0.90}
& 87.10 \up{0.17}
& 65.90 \up{1.52}
& 85.69 \down{0.13}
& 64.04 \up{0.23} \\

MMMLU
& 86.01 \up{0.67}
& 62.91 \up{1.49}
& 87.21 \up{0.72}
& 66.47 \up{1.99}
& 85.89 \up{0.51}
& 64.55 \up{1.53} \\

XCSQA
& 85.82 \up{1.19}
& 63.18 \up{3.41}
& 87.49 \up{2.06}
& 67.67 \up{5.54}
& 85.97 \up{0.56}
& 65.36 \up{2.36} \\

\bottomrule
\end{tabular}

\caption{
CLC scores on BLEND for Qwen2.5, Gemma3, and Aya after post-training on BMLAMA, MMMLU, or XCSQA with DCO, averaged over all tested country pairs. CosSim denotes cosine similarity between response embeddings from \texttt{BAAI/bge-m3}. Arrows indicate changes from Baseline.
}
\label{tab:consistency-main}
\end{table*}

Following \Cref{sec:geomalama-indomain}, we divide GEOMLAMA queries into culture-independent and culture-diverse subsets for each English--target language pair; see \Cref{tab:geomlama_category_coverage} for statistics. Although both subsets come from the same dataset and cover the same broad categories,\footnote{The categories include rules, policies, geography, customs, personal choices, and habits, following \citealp{yin-etal-2022-geomlama}.}
they are linearly separable using both
off-the-shelf lexical
representations and the hidden-state representations of the evaluated LLMs,
as shown in \Cref{fig:probe_scores}.\footnote{
Due to space constraints, we report results using concatenated features. For lexical representations, we present Word2Vec results; for hidden states, we present results from the final layer as a representative late-layer view. The qualitative patterns remain consistent for TF-IDF, other layers (e.g., middle layers), and the remaining feature constructions; see \Cref{app:full_results} for additional results.
}

These results suggest that the two subsets, beyond differences in gold-answer agreement, exhibit distinguishable distributions in both the input text and the model's representation spaces. This provides a plausible explanation for the limited transfer observed above: post-training on culture-independent questions may primarily affect representations associated with those questions, with weaker effects on culturally diverse ones.

\subsection{Extended Evaluation on BLEND}
\label{sec:extension_blend}

The GEOMLAMA results suggest that CLC improvements transfer only weakly to culturally diverse knowledge. 
If this pattern generalizes, it would be desirable in practice: models could improve cross-lingual consistency on culture-independent content without substantially altering culturally specific behavior. We therefore extend the evaluation to BLEND to examine whether this pattern holds in a more realistic setting with open-ended, culturally grounded generation.

We evaluate models post-trained on BMLAMA, MMMLU, and XCSQA with DCO, which shows the most stable cross-domain generalization,
using the seven English--non-English language pairs shared with BLEND. 
Following the original BLEND setup~\citep{myung2024blend}, we use greedy decoding and prepend a short country-agnostic instruction to encourage natural responses and reduce degenerate continuation or repetition~\citep{Holtzman2019TheCC,zekri2024large}. \looseness=-1 

\begin{table*}[!t]
\centering
\small
\setlength{\tabcolsep}{6pt}
\renewcommand{\arraystretch}{1.08}

\begin{tabular}{lcccccc}
\toprule
&
\multicolumn{2}{c}{Qwen2.5} &
\multicolumn{2}{c}{Gemma3} &
\multicolumn{2}{c}{Aya} \\
\cmidrule(lr){2-3}
\cmidrule(lr){4-5}
\cmidrule(lr){6-7}
Post-training Data
& Acc. EN & Acc. Non
& Acc. EN & Acc. Non
& Acc. EN & Acc. Non \\
\midrule

BMLAMA
& 76.50 \up{0.50}
& 43.51 \up{0.29}
& 73.64 \up{1.01}
& 42.50 \down{2.18}
& 79.60 \up{0.16}
& 48.25 \up{0.17} \\

MMMLU
& 76.50 \up{0.48}
& 48.30 \down{0.52}
& 73.64 \up{0.74}
& 44.48 \down{0.26}
& 79.60 \up{0.36}
& 48.90 \down{0.08} \\

XCSQA
& 76.50 \up{2.27}
& 55.40 \up{0.69}
& 73.64 \down{0.88}
& 48.55 \down{4.28}
& 79.60 \down{0.20}
& 52.56 \down{0.04} \\

\bottomrule
\end{tabular}

\caption{
Accuracy on BLEND for after post-training
on BMLAMA, MMMLU, or XCSQA with DCO, averaged separately over English (Acc. EN)
and non-English languages (Acc. Non). Arrows indicate changes from
Baseline.
}
\label{tab:acc-semb-main}
\end{table*}
The consistency results in \Cref{tab:consistency-main}, measured by BERTScore and cosine similarity (cf. \Cref{sec:evalSetups}), 
show modest but generally positive gains after DCO post-training. Embedding cosine similarity improves by $0.23$--$5.54$ across datasets and models, while BERTScore increases in all but one setting. The only exception is Aya post-trained on BMLAMA, where BERTScore decreases marginally by $0.13$ despite a $0.23$ increase in cosine similarity.

The accuracy results in \Cref{tab:acc-semb-main} are generally stable but more mixed. Across settings, accuracy may slightly improve in both English and non-English languages, improve in English while declining in non-English languages, or decrease in both. 
Decreases are more frequent for non-English languages, with the largest drop being $4.28$ for Gemma3 post-trained on XCSQA. 

Overall, DCO's semantic similarity gains indicate that the CLC improvements can extend from closed-form evaluation to open-ended generation. However, unlike on GEOMLAMA, improved CLC on BLEND is occasionally accompanied by reduced accuracy, especially for culturally grounded non-English responses. While this degradation is neither universal nor large in our evaluation, it suggests that open-ended generation may expose a trade-off between cross-lingual consistency and culturally appropriate behavior that is less visible in closed-form QA.

\section{Conclusion}


We present a unified evaluation of representative methods for cross-lingual consistency enhancement in multilingual language models across closed-form and open-ended question-answering tasks. Our results show substantial differences in reliability: inference-time intervention is brittle, DPO-based alignment is sensitive to question format and supervision quality, multilingual self-alignment yields modest gains under noisy preferences, and direct distribution alignment is the most robust overall. However, all evaluated methods show limited cross-domain generalization, indicating that strong in-domain gains do not necessarily translate into broader multilingual consistency.

We further examine whether CLC enhancement compromises culturally diverse knowledge. On GEOMLAMA, post-training methods do not systematically degrade accuracy on culturally diverse questions
and our finding that the prompts are linearly separable at both the lexical and representation levels offers a plausible explanation for this limited interference.
On BLEND, however, improved cross-lingual similarity is occasionally accompanied by lower accuracy, particularly for non-English responses. 
This reveals a more nuanced trade-off: 
while CLC enhancement can improve cross-lingual semantic alignment, these gains may come at the cost of answer accuracy, especially for culturally grounded knowledge from contexts that are less represented in English-centric data.


Overall, our findings highlight the need to evaluate CLC enhancement beyond in-domain consistency, with attention to cross-domain robustness and culturally grounded settings. Future work should develop methods that generalize more reliably across tasks while preserving culture-specific behavior where such differences are warranted.




\section*{Limitations}


One limitation is the use of semantic similarity as a proxy for cross-lingual consistency on BLEND benchmark. While semantic similarity provides a practical and scalable measure of response-level alignment, it may not fully capture differences in culturally grounded knowledge across languages. 

Besides, our analysis of culturally diverse knowledge is based on GEOMLAMA and BLEND. While these benchmarks provide complementary closed-form and open-ended settings and are sufficient for our controlled study, broader benchmark coverage and larger-scale human annotation would be valuable for assessing the impact of CLC-enhancement methods on culturally diverse knowledge in real-world multilingual systems.

Finally, while our separability analysis reveals a clear distinction between culture-independent and culturally diverse prompts, this finding should not be interpreted as direct evidence that the models explicitly encode cultural (in)dependence. In particular, representation-level separation may partly inherit systematic differences already present in the inputs, such as entity distributions, translation styles, prompt-template variation, or language-specific lexical cues. Disentangling culturally grounded representations from these potential dataset- and surface-level confounders would require more controlled interventions and causal analyses, which we leave to future work.

\section*{Acknowledge}
The authors have received funding from the Dutch Research Council (NWO): JQ is supported by NWA-ORC project LESSEN (grant nr. NWA.1389.20.183). AB is supported by the above as well as NWO Talent Programme (VI.Vidi.221C.009). This research was also supported by DFG (grant SCHU 2246/14-1).

This work used the Dutch national e-infrastructure with the support of the SURF Cooperative using grant no. EINF-17648. 

\bibliography{custom, anthology-1, anthology-2}
\bibliographystyle{acl_natbib}


\onecolumn

\appendix
\crefalias{section}{appendix} 

\section{Additional Details}
\label{app:dataset_examples}

\subsection{Statistic of Evaluation Datasets}

The languages covered by this study is shown in \Cref{tab:language_coverage_new}. \Cref{tab:dataset_examples} presents representative examples from the evaluation datasets used in our experiments. 

\begin{table*}[!h]
\centering
\setlength{\tabcolsep}{2.0pt}
\resizebox{\textwidth}{!}{%
\begin{tabular}{l*{32}{c}}
\toprule
\textbf{Dataset}
& \textsc{am} & \textsc{ar} & \textsc{as} & \textsc{az} & \textsc{bn} & \textsc{ca}
& \textsc{de} & \textsc{el} & \textsc{en} & \textsc{es} & \textsc{fa} & \textsc{fr}
& \textsc{ha} & \textsc{he} & \textsc{hi} & \textsc{hu} & \textsc{id} & \textsc{it}
& \textsc{ja} & \textsc{ko} & \textsc{nl} & \textsc{pl} & \textsc{pt} & \textsc{ru}
& \textsc{su} & \textsc{sw} & \textsc{tr} & \textsc{uk} & \textsc{ur} & \textsc{vi}
& \textsc{yo} & \textsc{zh} \\
\midrule
BMLAMA
& -- & \cmark & -- & -- & -- & \cmark & -- & \cmark & \cmark & \cmark
& \cmark & \cmark & -- & \cmark & -- & \cmark
& -- & -- & \cmark & \cmark & \cmark & -- & -- & \cmark
& -- & -- & \cmark & \cmark & -- & \cmark & -- & \cmark \\

MMMLU
& -- & \cmark & -- & -- & \cmark & -- & \cmark & -- & \cmark & \cmark
& -- & \cmark & -- & -- & \cmark & --
& \cmark & \cmark & \cmark & \cmark & -- & -- & \cmark & --
& -- & \cmark & -- & -- & -- & -- & \cmark & \cmark \\

XCSQA
& -- & \cmark & -- & -- & -- & -- & \cmark & -- & \cmark & \cmark
& -- & \cmark & -- & -- & \cmark & --
& -- & \cmark & \cmark & -- & \cmark & \cmark & \cmark & \cmark
& -- & \cmark & -- & -- & \cmark & \cmark & -- & \cmark \\

GEOMLAMA
& -- & -- & -- & -- & -- & -- & -- & -- & \cmark & --
& \cmark & -- & -- & -- & \cmark & --
& -- & -- & -- & -- & -- & -- & -- & --
& -- & \cmark & -- & -- & -- & -- & -- & \cmark \\

BLEND
& \cmark & \cmark & \cmark & \cmark & -- & -- & -- & \cmark & \cmark & \cmark
& \cmark & -- & \cmark & -- & -- & --
& \cmark & -- & -- & \cmark & -- & -- & -- & --
& \cmark & -- & -- & -- & -- & -- & -- & \cmark \\
\bottomrule
\end{tabular}%
}
\caption{Detailed language coverage of the evaluated datasets. Overall, the datasets span 32 languages.}
\label{tab:language_coverage_new}
\end{table*}

\begin{table*}[!h]
\centering
\setlength{\tabcolsep}{2.8pt}
\resizebox{\textwidth}{!}{
\begin{tabular}{p{2.3cm}p{6.2cm}p{6.1cm}p{2.8cm}p{2.3cm}}
\toprule
\textbf{Dataset} & \textbf{Example Input} & \textbf{Candidates / Answer Space} & \textbf{Culture/Region} & \textbf{Gold Answer} \\
\midrule
BMLAMA &
\textit{Charles II of Spain was born in \_\_\_.} &
\textit{Toronto; London; Belgrade; Manchester; Naples; Brooklyn; Vienna; Istanbul; Geneva; Madrid} & - &
\textit{Madrid} \\
\midrule
MMMLU &
\textit{Find the degree of the field extension \(Q(\sqrt{2}, \sqrt{3}, \sqrt{18})\) over \(Q\).} &
\textit{A: 0; B: 4; C: 2; D: 6} & - & 
\textit{B} \\
\midrule
XCSQA &
\textit{What will happen to your knowledge with more learning?} &
\textit{A: headaches; B: bigger brain; C: education; D: growth; E: knowing more} & - & 
\textit{D} \\
\midrule
GEOMLAMA &
\textit{People normally shower in the \_\_\_.} &
\textit{morning; noon; afternoon; evening} &
\textit{China} &
\textit{evening} \\
\midrule
BLEND &
\textit{What is a common school cafeteria food?} & - & \textit{South Korea} & \textit{kimchi; rice} \\
\bottomrule
\end{tabular}
}
\caption{Examples from the evaluation datasets. For readability, all examples are shown in English. During evaluation, the LLM is provided with the corresponding parallel versions and evaluated with either culture-independent or culture-diverse gold answers.}
\label{tab:dataset_examples}
\end{table*}

\subsection{Definition of RankC: Ranking-based Crosslingual Consistency}
\label{app:rankc}
RankC \citep{qi-etal-2023-cross} measures cross-lingual consistency by comparing the rankings of candidate responses induced by semantically equivalent prompts, without conflating consistency with probing accuracy.
Given two parallel prompts $(\bxone, \bxtwo)$ and a shared set of $M$ candidate responses, we sort the candidates in descending order of their average likelihood under each prompt, obtaining
$\candone^{1}, \candone^{2}, \ldots, \candone^{M}$ for language $L_1$ and
$\candtwo^{1}, \candtwo^{2}, \ldots, \candtwo^{M}$ for language $L_2$.
For each cutoff $j$, RankC computes the overlap between the two top-$j$ candidate sets:
\begin{equation}
\text{P@}j =
\frac{
\left|
\{\candone^{1}, \ldots, \candone^{j}\}
\cap
\{\candtwo^{1}, \ldots, \candtwo^{j}\}
\right|
}{j}.
\end{equation}
To place greater emphasis on agreement among highly ranked candidates, each $\text{P@}j$ is assigned a normalized exponentially decaying weight,
\begin{equation}
w_j =
\frac{\exp(M-j)}
{\sum_{k=1}^{M}\exp(M-k)}.
\end{equation}
The final consistency score is the weighted average over all ranking depths:
\begin{equation}
\mathrm{RankC}(\bxone, \bxtwo)
=
\sum_{j=1}^{M} w_j\,\text{P@}j.
\end{equation}

\section{Complementary Results}
\label{app:full_results}

\subsection{Full Results of CLC Enhancement Methods}
The full results of the CLC enhancement methods are reported in \Cref{tab:joint_all_results}.
\begin{table*}[!h]
\centering
\begingroup

\resizebox{\textwidth}{!}{%
\begin{tabular}{@{}llcrrrrrrrrrrrrrrrrrr@{}}
\toprule
Model & Method & M. & Avg. & \textsc{en} & \textsc{fr} & \textsc{nl} & \textsc{es} & \textsc{ru} & \textsc{ja} & \textsc{zh} & \textsc{ko} & \textsc{vi} & \textsc{el} & \textsc{hu} & \textsc{he} & \textsc{tr} & \textsc{ca} & \textsc{ar} & \textsc{uk} & \textsc{fa} \\
\midrule 
\\ [-16.5pt] \rowcolor{black!8} \multicolumn{21}{c}{\textbf{Consistency / Accuracy results on BMLAMA}} \\[-1.5pt]
\midrule
\multirow{8}{*}{\texttt{Qwen2.5}}
& \multirow{2}{*}{Base}
& \cellcolor{gray!10}\textsc{C}
& 38.97 & -- & 44.42 & 48.32 & 46.58 & 41.84 & 39.91 & 40.86 & 36.42 & 45.10 & 30.80 & 30.28 & 30.64 & 35.29 & 37.78 & 39.82 & 40.12 & 35.37 \\
&
& \cellcolor{gray!18}\textsc{A}
& 36.86 & 62.22 & 36.55 & 43.64 & 42.13 & 40.18 & 36.72 & 36.05 & 34.43 & 38.45 & 27.18 & 27.62 & 25.67 & 32.81 & 33.93 & 38.45 & 37.50 & 33.09 \\

& \multirow{2}{*}{+CALM}
& \cellcolor{gray!10}\textsc{C}
& +2.27 & -- & +2.41 & +2.29 & +1.67 & +2.40 & +1.87 & +0.95 & +1.92 & +2.34 & +3.35 & +2.69 & +1.84 & +2.66 & +2.68 & +2.40 & +2.70 & +2.23 \\
&
& \cellcolor{gray!18}\textsc{A}
& +2.07 & +0.89 & +2.96 & +2.12 & +1.73 & +1.62 & +1.84 & +0.95 & +1.51 & +2.06 & +3.18 & +3.02 & +2.23 & +1.34 & +2.57 & +2.57 & +2.57 & +2.01 \\

& \multirow{2}{*}{+EN-A}
& \cellcolor{gray!10}\textsc{C}
& +20.99 & -- & +23.22 & +19.80 & +21.54 & +22.59 & +24.34 & +20.36 & +20.85 & +19.33 & +16.83 & +19.50 & +20.76 & +21.90 & +25.11 & +19.06 & +19.54 & +21.10 \\
&
& \cellcolor{gray!18}\textsc{A}
& +18.14 & +3.24 & +24.16 & +15.57 & +20.59 & +18.47 & +21.65 & +22.10 & +21.65 & +17.13 & +14.39 & +17.92 & +19.64 & +19.53 & +23.27 & +14.95 & +16.85 & +17.19 \\

& \multirow{2}{*}{+DCO}
& \cellcolor{gray!10}\textsc{C}
& +15.64 & -- & +19.58 & +18.59 & +16.32 & +17.78 & +17.35 & +15.32 & +14.34 & +16.22 & +13.29 & +13.11 & +13.94 & +12.92 & +18.55 & +13.36 & +16.81 & +12.71 \\
&
& \cellcolor{gray!18}\textsc{A}
& +14.99 & +6.53 & +21.82 & +16.91 & +16.41 & +15.57 & +17.69 & +17.30 & +15.12 & +16.91 & +11.38 & +12.84 & +14.56 & +11.33 & +18.13 & +12.39 & +17.02 & +12.95 \\

\cmidrule(lr){1-21}

\multirow{8}{*}{\texttt{Gemma3}}
& \multirow{2}{*}{Base}
& \cellcolor{gray!10}\textsc{C}
& 35.19 & -- & 35.16 & 46.32 & 39.10 & 37.30 & 30.76 & 26.82 & 31.11 & 44.94 & 37.00 & 27.91 & 34.83 & 33.15 & 35.26 & 33.60 & 39.37 & 30.36 \\
&
& \cellcolor{gray!18}\textsc{A}
& 31.41 & 63.62 & 28.01 & 40.18 & 33.31 & 30.25 & 23.88 & 22.38 & 27.73 & 38.50 & 29.80 & 21.09 & 29.07 & 27.79 & 30.80 & 29.46 & 34.15 & 23.94 \\

& \multirow{2}{*}{+CALM}
& \cellcolor{gray!10}\textsc{C}
& +0.20 & -- & +0.38 & +0.21 & +0.29 & -0.10 & +0.08 & +0.31 & -0.03 & +0.38 & +0.10 & +0.37 & +0.26 & -0.17 & +0.43 & +0.40 & -0.01 & +0.25 \\
&
& \cellcolor{gray!18}\textsc{A}
& +0.25 & -0.34 & +0.23 & +1.00 & +0.51 & +0.55 & +0.06 & +0.28 & -0.05 & +0.34 & +0.11 & +0.12 & +0.12 & -0.06 & +0.62 & +0.34 & +0.23 & +0.22 \\

& \multirow{2}{*}{+EN-A}
& \cellcolor{gray!10}\textsc{C}
& +2.14 & -- & +2.64 & +2.18 & +2.84 & +2.31 & +1.10 & +1.69 & +1.19 & +3.71 & +2.55 & +2.51 & +1.94 & +0.49 & +2.01 & +2.43 & +3.03 & +1.66 \\
&
& \cellcolor{gray!18}\textsc{A}
& +1.84 & +0.16 & +2.29 & +2.40 & +2.91 & +2.73 & +1.06 & +2.06 & +1.34 & +2.79 & +2.29 & +0.95 & +2.24 & -0.33 & +2.46 & +2.68 & +1.45 & +1.79 \\

& \multirow{2}{*}{+DCO}
& \cellcolor{gray!10}\textsc{C}
& +20.37 & -- & +26.92 & +19.01 & +22.19 & +22.51 & +23.87 & +19.01 & +22.71 & +21.52 & +16.58 & +23.11 & +15.77 & +21.17 & +20.47 & +14.10 & +21.29 & +15.65 \\
&
& \cellcolor{gray!18}\textsc{A}
& +19.58 & +3.90 & +26.96 & +17.52 & +21.49 & +23.32 & +25.62 & +21.26 & +23.22 & +17.36 & +18.47 & +23.89 & +14.23 & +21.82 & +18.81 & +15.52 & +21.43 & +18.08 \\

\cmidrule(lr){1-21}

\multirow{8}{*}{\texttt{Aya}}
& \multirow{2}{*}{Base}
& \cellcolor{gray!10}\textsc{C}
& 41.89 & -- & 50.73 & 51.41 & 58.04 & 39.67 & 39.71 & 38.43 & 36.55 & 55.06 & 34.62 & 29.14 & 36.88 & 41.92 & 36.98 & 40.15 & 44.64 & 36.38 \\
&
& \cellcolor{gray!18}\textsc{A}
& 39.51 & 66.85 & 45.09 & 47.15 & 52.57 & 35.21 & 36.22 & 37.67 & 33.87 & 49.44 & 32.81 & 22.27 & 33.26 & 35.88 & 32.70 & 37.56 & 39.96 & 33.20 \\

& \multirow{2}{*}{+CALM}
& \cellcolor{gray!10}\textsc{C}
& +0.37 & -- & +0.43 & +0.26 & +0.60 & +0.36 & +0.32 & +0.23 & +0.39 & +0.08 & +0.33 & +0.02 & +0.27 & +0.63 & +0.74 & +0.50 & +0.30 & +0.42 \\
&
& \cellcolor{gray!18}\textsc{A}
& +0.34 & -0.05 & +0.61 & +0.23 & +0.44 & +0.28 & +0.33 & +0.11 & +0.45 & +0.39 & +0.28 & +0.16 & +0.39 & +0.62 & +0.73 & +0.55 & +0.05 & +0.17 \\

& \multirow{2}{*}{+EN-A}
& \cellcolor{gray!10}\textsc{C}
& +11.57 & -- & +14.91 & +12.53 & +8.09 & +16.51 & +12.70 & +12.95 & +11.83 & +6.12 & +14.52 & +6.45 & +11.82 & +12.04 & +12.26 & +8.60 & +11.65 & +12.17 \\
&
& \cellcolor{gray!18}\textsc{A}
& +9.80 & +0.90 & +13.84 & +11.72 & +8.20 & +16.69 & +11.94 & +10.49 & +10.55 & +5.75 & +10.88 & +2.73 & +7.25 & +12.72 & +12.11 & +7.81 & +11.88 & +11.22 \\

& \multirow{2}{*}{+DCO}
& \cellcolor{gray!10}\textsc{C}
& +12.86 & -- & +17.82 & +14.62 & +14.95 & +16.11 & +14.27 & +12.91 & +11.42 & +14.83 & +11.48 & +5.65 & +11.15 & +12.17 & +12.48 & +10.58 & +13.30 & +12.04 \\
&
& \cellcolor{gray!18}\textsc{A}
& +12.64 & +3.85 & +16.80 & +14.46 & +11.94 & +18.70 & +15.06 & +13.73 & +13.06 & +12.61 & +12.34 & +4.40 & +12.05 & +13.90 & +12.05 & +11.16 & +14.56 & +14.29 \\

\bottomrule
\end{tabular}%
}


\resizebox{\textwidth}{!}{%
\begin{tabular}{@{}llcrrrrrrrrrrrrrrrr@{}}
\\ [-16.5pt]
\toprule
Model & Method & M. & Avg. & \textsc{en} & \textsc{ar} & \textsc{de} & \textsc{es} & \textsc{fr} & \textsc{hi} & \textsc{id} & \textsc{it} & \textsc{ja} & \textsc{ko} & \textsc{pt} & \textsc{sw} & \textsc{yo} & \textsc{zh} & \textsc{bn} \\
\midrule
\\ [-16.5pt] \rowcolor{black!8} \multicolumn{19}{c}{\textbf{Consistency / Accuracy results on MMMLU}} \\[-1.5pt]
\midrule
\multirow{8}{*}{\texttt{Qwen2.5}}
& \multirow{2}{*}{Base}
& \cellcolor{gray!10}\textsc{C}
& 66.44 & -- & 68.36 & 72.27 & 77.51 & 74.40 & 56.98 & 70.22 & 74.91 & 70.69 & 68.53 & 70.29 & 47.81 & 45.99 & 75.39 & 56.82 \\
&
& \cellcolor{gray!18}\textsc{A}
& 55.69 & 70.89 & 55.39 & 61.12 & 65.22 & 64.39 & 45.00 & 59.84 & 63.51 & 59.49 & 58.01 & 58.50 & 33.27 & 31.08 & 66.07 & 43.64 \\

& \multirow{2}{*}{+CALM}
& \cellcolor{gray!10}\textsc{C}
& +2.36 & -- & +1.67 & +1.08 & +1.50 & +3.12 & +4.08 & +3.70 & +2.89 & +2.07 & +1.68 & +7.55 & +0.55 & +0.50 & +1.12 & +1.58 \\
&
& \cellcolor{gray!18}\textsc{A}
& +1.21 & +0.20 & +0.70 & +0.34 & +0.48 & +0.86 & +2.19 & +1.22 & +0.73 & +1.50 & +0.69 & +6.11 & +1.21 & +0.74 & -0.28 & +1.44 \\

& \multirow{2}{*}{+EN-A}
& \cellcolor{gray!10}\textsc{C}
& +1.18 & -- & +0.98 & +1.77 & +1.26 & +2.41 & +1.03 & +3.06 & +2.36 & +0.92 & +0.57 & +0.69 & +1.24 & -0.06 & -0.03 & +0.26 \\
&
& \cellcolor{gray!18}\textsc{A}
& -0.33 & +0.06 & -0.56 & -0.67 & -0.65 & -0.12 & -0.41 & -0.32 & -0.01 & +0.06 & -0.36 & -0.79 & -0.29 & +0.42 & -1.38 & +0.12 \\

& \multirow{2}{*}{+DCO}
& \cellcolor{gray!10}\textsc{C}
& +8.90 & -- & +8.35 & +9.63 & +7.92 & +9.66 & +11.33 & +11.22 & +9.43 & +8.55 & +8.37 & +14.44 & +5.87 & +3.16 & +7.39 & +9.35 \\
&
& \cellcolor{gray!18}\textsc{A}
& +2.45 & -0.20 & +2.70 & +2.59 & +1.22 & +1.18 & +5.37 & +2.85 & +2.03 & +2.87 & +1.82 & +8.21 & +2.14 & +0.78 & -0.34 & +3.60 \\

\cmidrule(lr){1-19}

\multirow{8}{*}{\texttt{Gemma3}}
& \multirow{2}{*}{Base}
& \cellcolor{gray!10}\textsc{C}
& 65.27 & -- & 65.15 & 68.13 & 73.20 & 69.55 & 63.16 & 68.16 & 71.31 & 65.20 & 65.11 & 72.49 & 57.66 & 46.34 & 68.08 & 60.26 \\
&
& \cellcolor{gray!18}\textsc{A}
& 49.11 & 57.63 & 47.85 & 52.57 & 54.36 & 53.11 & 48.04 & 51.72 & 53.44 & 48.61 & 48.95 & 52.52 & 40.20 & 32.11 & 50.58 & 44.99 \\

& \multirow{2}{*}{+CALM}
& \cellcolor{gray!10}\textsc{C}
& +1.10 & -- & +0.90 & +1.33 & +0.45 & +1.54 & +2.34 & +1.79 & +0.18 & +2.03 & -0.20 & -0.46 & -0.09 & +2.15 & +0.60 & +2.78 \\
&
& \cellcolor{gray!18}\textsc{A}
& -0.34 & +0.28 & -0.79 & -0.70 & -0.69 & -0.33 & -0.58 & -0.15 & -0.60 & -0.01 & -0.55 & -0.01 & -0.26 & +0.27 & -0.36 & -0.62 \\

& \multirow{2}{*}{+EN-A}
& \cellcolor{gray!10}\textsc{C}
& +5.32 & -- & +5.78 & +7.23 & +4.72 & +4.73 & +5.59 & +4.95 & +5.14 & +5.74 & +4.25 & +4.69 & +4.49 & +8.92 & +3.76 & +4.46 \\
&
& \cellcolor{gray!18}\textsc{A}
& -2.44 & -1.96 & -3.12 & -2.60 & -2.83 & -2.11 & -2.33 & -2.40 & -2.56 & -2.20 & -2.35 & -2.51 & -2.27 & -3.20 & -2.58 & -1.61 \\

& \multirow{2}{*}{+DCO}
& \cellcolor{gray!10}\textsc{C}
& +11.61 & -- & +11.33 & +13.54 & +10.10 & +12.80 & +13.65 & +13.01 & +11.69 & +11.44 & +10.67 & +10.50 & +11.23 & +8.73 & +10.48 & +13.40 \\
&
& \cellcolor{gray!18}\textsc{A}
& +0.83 & +0.14 & +0.17 & +0.66 & +0.19 & +0.22 & +0.79 & +0.81 & +0.34 & +0.82 & +0.68 & +1.33 & +2.50 & +1.01 & +0.78 & +2.04 \\

\cmidrule(lr){1-19}

\multirow{8}{*}{\texttt{Aya}}
& \multirow{2}{*}{Base}
& \cellcolor{gray!10}\textsc{C}
& 66.63 & -- & 69.59 & 74.31 & 77.36 & 76.02 & 65.09 & 73.43 & 76.28 & 69.07 & 68.52 & 71.12 & 46.36 & 45.03 & 69.88 & 50.72 \\
&
& \cellcolor{gray!18}\textsc{A}
& 49.25 & 59.79 & 50.83 & 54.14 & 56.36 & 55.76 & 46.59 & 53.12 & 54.63 & 52.02 & 51.40 & 55.03 & 31.88 & 30.43 & 52.93 & 33.90 \\

& \multirow{2}{*}{+CALM}
& \cellcolor{gray!10}\textsc{C}
& +2.18 & -- & +1.67 & +1.01 & +1.65 & +1.83 & +1.08 & +1.28 & +1.72 & +2.50 & +2.05 & +2.24 & +4.40 & +3.72 & +2.00 & +3.34 \\
&
& \cellcolor{gray!18}\textsc{A}
& -0.57 & -0.43 & -1.21 & -0.79 & -0.91 & -0.79 & -0.69 & -1.12 & -0.59 & -0.47 & -0.95 & -0.22 & +0.28 & -0.20 & -0.02 & -0.50 \\

& \multirow{2}{*}{+EN-A}
& \cellcolor{gray!10}\textsc{C}
& +0.22 & -- & -0.12 & -0.68 & +0.15 & -0.95 & +0.23 & -0.54 & +0.23 & +0.41 & -0.28 & +4.03 & +0.02 & +1.60 & -0.01 & -1.04 \\
&
& \cellcolor{gray!18}\textsc{A}
& -0.61 & +0.36 & -0.77 & -0.90 & -0.80 & -0.68 & -0.92 & -0.77 & -0.36 & -0.67 & -0.56 & +0.07 & -0.62 & -0.59 & -1.12 & -0.88 \\

& \multirow{2}{*}{+DCO}
& \cellcolor{gray!10}\textsc{C}
& +8.96 & -- & +8.24 & +8.46 & +7.06 & +7.89 & +8.46 & +8.25 & +7.79 & +8.99 & +9.10 & +12.73 & +10.56 & +9.85 & +8.69 & +9.43 \\
&
& \cellcolor{gray!18}\textsc{A}
& +0.94 & +0.68 & +0.61 & +1.35 & +0.29 & +0.65 & +1.48 & +0.98 & +1.68 & +0.92 & +1.18 & +1.10 & +0.95 & +0.48 & +0.62 & +1.09 \\

\bottomrule
\end{tabular}%
}

\resizebox{\textwidth}{!}{%
\begin{tabular}{@{}llcrrrrrrrrrrrrrrrrr@{}}
\\[-16.5pt]
\toprule
Model & Method & M. & Avg. & \textsc{en} & \textsc{zh} & \textsc{de} & \textsc{es} & \textsc{fr} & \textsc{it} & \textsc{ja} & \textsc{nl} & \textsc{pl} & \textsc{pt} & \textsc{ru} & \textsc{ar} & \textsc{vi} & \textsc{hi} & \textsc{sw} & \textsc{ur} \\
\midrule
\\ [-16.5pt] \rowcolor{black!8} \multicolumn{20}{c}{\textbf{Consistency / Accuracy results on XCSQA}} \\[-1.5pt]
\midrule
\multirow{8}{*}{\texttt{Qwen2.5}}
& \multirow{2}{*}{Base}
& \cellcolor{gray!10}\textsc{C}
& 60.34 & -- & 63.75 & 66.62 & 69.84 & 66.11 & 67.33 & 59.57 & 62.42 & 63.06 & 59.57 & 63.67 & 62.20 & 66.06 & 49.43 & 35.79 & 49.66 \\
&
& \cellcolor{gray!18}\textsc{A}
& 53.34 & 84.00 & 58.50 & 57.00 & 62.00 & 56.50 & 60.50 & 51.00 & 55.00 & 53.00 & 45.00 & 53.00 & 51.00 & 61.00 & 39.00 & 27.00 & 40.00 \\

& \multirow{2}{*}{+CALM}
& \cellcolor{gray!10}\textsc{C}
& +0.56 & -- & -1.00 & -1.16 & +0.87 & +0.55 & -1.65 & -0.49 & -1.24 & +0.05 & +11.01 & -1.33 & +1.87 & -0.68 & +1.76 & -0.71 & +0.60 \\
&
& \cellcolor{gray!18}\textsc{A}
& +0.00 & -4.50 & -3.50 & -2.00 & +0.00 & -0.50 & -2.50 & +0.50 & -1.00 & +3.00 & +13.00 & -0.50 & +2.50 & +1.00 & -1.00 & -5.00 & +0.50 \\

& \multirow{2}{*}{+EN-A}
& \cellcolor{gray!10}\textsc{C}
& -1.38 & -- & -1.60 & -3.06 & -1.92 & -1.68 & -6.18 & -6.53 & -0.07 & -4.75 & +4.63 & -6.98 & +3.29 & -3.81 & +1.94 & +6.80 & -0.74 \\
&
& \cellcolor{gray!18}\textsc{A}
& -13.38 & -23.00 & -14.00 & -15.50 & -22.00 & -16.00 & -18.50 & -17.50 & -15.50 & -12.00 & +6.50 & -13.50 & -10.50 & -22.00 & -6.00 & -7.50 & -7.00 \\

& \multirow{2}{*}{+DCO}
& \cellcolor{gray!10}\textsc{C}
& +10.11 & -- & +9.39 & +9.15 & +10.99 & +10.78 & +10.61 & +9.46 & +15.62 & +9.35 & +20.66 & +8.52 & +10.38 & +10.36 & +7.23 & +2.36 & +6.76 \\
&
& \cellcolor{gray!18}\textsc{A}
& +5.16 & -4.50 & +4.50 & +5.00 & +1.00 & +6.00 & +3.00 & +6.00 & +8.00 & +7.50 & +20.00 & +3.00 & +5.50 & +4.50 & +6.50 & +0.50 & +6.00 \\

\cmidrule(lr){1-20}

\multirow{8}{*}{\texttt{Gemma3}}
& \multirow{2}{*}{Base}
& \cellcolor{gray!10}\textsc{C}
& 58.41 & -- & 60.30 & 59.59 & 66.80 & 63.24 & 64.67 & 55.32 & 61.36 & 57.14 & 64.10 & 63.02 & 56.52 & 54.60 & 55.05 & 42.80 & 51.61 \\
&
& \cellcolor{gray!18}\textsc{A}
& 45.91 & 63.50 & 47.00 & 49.00 & 50.00 & 49.00 & 48.50 & 39.50 & 49.00 & 46.50 & 47.50 & 45.50 & 42.00 & 47.50 & 42.00 & 31.00 & 37.00 \\

& \multirow{2}{*}{+CALM}
& \cellcolor{gray!10}\textsc{C}
& +0.54 & -- & +0.90 & +0.54 & +0.67 & +1.21 & +0.31 & +0.35 & +0.12 & +0.06 & -0.75 & -0.25 & +1.53 & +0.56 & +1.48 & +0.06 & +1.36 \\
&
& \cellcolor{gray!18}\textsc{A}
& -0.38 & +0.00 & +0.00 & -1.50 & +0.00 & +0.00 & -0.50 & +0.50 & -1.00 & -0.50 & -0.50 & -1.50 & +1.00 & -0.50 & -0.50 & -1.00 & +0.00 \\

& \multirow{2}{*}{+EN-A}
& \cellcolor{gray!10}\textsc{C}
& +0.88 & -- & +1.24 & +2.08 & -1.27 & +4.22 & +0.46 & -0.92 & -0.07 & +1.60 & +0.66 & +1.53 & -0.19 & +3.54 & -2.53 & +0.68 & +2.17 \\
&
& \cellcolor{gray!18}\textsc{A}
& -1.47 & +0.00 & +0.50 & -2.00 & -2.50 & +0.00 & -4.00 & -2.50 & +0.50 & -3.50 & -1.50 & -1.00 & +0.50 & +1.50 & -3.00 & -3.50 & -3.00 \\

& \multirow{2}{*}{+DCO}
& \cellcolor{gray!10}\textsc{C}
& +15.83 & -- & +15.97 & +19.70 & +14.33 & +18.35 & +14.11 & +15.57 & +15.08 & +14.90 & +15.45 & +14.66 & +16.76 & +19.89 & +14.86 & +8.88 & +18.99 \\
&
& \cellcolor{gray!18}\textsc{A}
& +5.81 & +0.50 & +6.00 & +7.50 & +6.00 & +8.50 & +6.00 & +10.00 & +6.00 & +2.00 & +5.50 & +7.50 & +8.50 & +4.50 & +3.00 & +4.00 & +7.50 \\

\cmidrule(lr){1-20}

\multirow{8}{*}{\texttt{Aya}}
& \multirow{2}{*}{Base}
& \cellcolor{gray!10}\textsc{C}
& 62.57 & -- & 67.26 & 69.43 & 73.27 & 71.04 & 69.63 & 56.74 & 67.69 & 64.82 & 65.41 & 64.85 & 64.43 & 66.08 & 58.57 & 34.54 & 44.83 \\
&
& \cellcolor{gray!18}\textsc{A}
& 55.91 & 78.00 & 61.50 & 59.00 & 65.00 & 58.00 & 63.00 & 49.50 & 60.50 & 57.50 & 60.00 & 58.50 & 55.00 & 60.00 & 50.00 & 23.00 & 36.00 \\

& \multirow{2}{*}{+CALM}
& \cellcolor{gray!10}\textsc{C}
& +0.05 & -- & -0.56 & +0.46 & +0.51 & -0.25 & +0.77 & +1.30 & +1.10 & -0.42 & -0.34 & -0.11 & +0.04 & -0.15 & +0.05 & +0.50 & -2.08 \\
&
& \cellcolor{gray!18}\textsc{A}
& -0.19 & +0.50 & +0.00 & -0.50 & -0.50 & -1.00 & -1.00 & +1.50 & +0.00 & +0.00 & +0.50 & +0.00 & -0.50 & -1.00 & +0.50 & +1.00 & -2.50 \\

& \multirow{2}{*}{+EN-A}
& \cellcolor{gray!10}\textsc{C}
& -0.09 & -- & -2.23 & +0.10 & -1.36 & -1.10 & +0.59 & +0.59 & +2.73 & -0.67 & +4.04 & -0.08 & -0.36 & -2.20 & -1.91 & +3.08 & -2.52 \\
&
& \cellcolor{gray!18}\textsc{A}
& -1.09 & -2.00 & -2.50 & -2.50 & +0.50 & -4.00 & -2.50 & -0.50 & -1.00 & -0.50 & +2.00 & -1.00 & +1.50 & -3.50 & -2.50 & +4.00 & -3.00 \\

& \multirow{2}{*}{+DCO}
& \cellcolor{gray!10}\textsc{C}
& +7.49 & -- & +5.96 & +4.44 & +7.51 & +5.74 & +8.55 & +9.35 & +7.92 & +5.55 & +14.34 & +6.56 & +8.23 & +8.80 & +9.70 & +1.31 & +8.41 \\
&
& \cellcolor{gray!18}\textsc{A}
& +2.56 & +1.00 & +4.00 & -1.00 & +1.50 & +2.50 & +3.00 & +6.00 & +0.00 & +2.50 & +6.50 & -0.50 & +2.00 & +2.50 & +3.50 & +2.00 & +5.50 \\

\bottomrule
\end{tabular}%
}

\caption{
Joint-language post-training results on BMLAMA, MMMLU, and XCSQA. M. denotes the evaluation metric.
\protect\colorbox{gray!10}{\textsc{C}} and
\protect\colorbox{gray!18}{\textsc{A}} denote consistency and accuracy scores, respectively.
Baseline rows report absolute scores, while post-training rows report the changes
relative to the corresponding baseline.
EN-A abbreviates EN-Align.
}
\label{tab:joint_all_results}

\endgroup
\end{table*}

\subsection{Full Results of Cross-domain Generalization}
\label{app:full-cross-domain}
\begin{table*}[!h]
\centering
\begingroup
\scriptsize
\setlength{\tabcolsep}{1.8pt}
\resizebox{0.9\textwidth}{!}{%
\begin{tabular}{@{}llcrrrrrrrrrrrrrrrrrrrrr@{}}
\toprule
\multirow{2}{*}{Model} & \multirow{2}{*}{Method} & \multirow{2}{*}{M.}
& \multicolumn{8}{c}{Train on BMLAMA, Test on MMMLU}
& \multicolumn{12}{c}{Train on XCSQA, Test on MMMLU} \\
\cmidrule(lr){4-11} \cmidrule(lr){12-23}
& & & Avg. & \textsc{en} & \textsc{fr} & \textsc{es} & \textsc{ar} & \textsc{ja} & \textsc{zh} & \textsc{ko}
& Avg. & \textsc{en} & \textsc{zh} & \textsc{de} & \textsc{es} & \textsc{fr} & \textsc{it} & \textsc{ja} & \textsc{pt} & \textsc{ar} & \textsc{hi} & \textsc{sw} \\
\midrule
\multirow{8}{*}{\texttt{Qwen2.5}}
& \multirow{2}{*}{Base}
& \cellcolor{gray!10}\textsc{C}
& 72.48 & -- & 74.40 & 77.51 & 68.36 & 70.69 & 75.39 & 68.53 & 68.86 & -- & 75.39 & 72.27 & 77.51 & 74.40 & 74.91 & 70.69 & 70.29 & 68.36 & 56.98 & 47.81 \\
&
& \cellcolor{gray!18}\textsc{A}
& 62.78 & 70.89 & 64.39 & 65.22 & 55.39 & 59.49 & 66.07 & 58.01 & 58.44 & 70.89 & 66.07 & 61.12 & 65.22 & 64.39 & 63.51 & 59.49 & 58.50 & 55.39 & 45.00 & 33.27 \\
& \multirow{2}{*}{+EN-A}
& \cellcolor{gray!10}\textsc{C}
& -0.15 & -- & +0.09 & -0.29 & -0.19 & -0.16 & +0.05 & -0.42 & -0.41 & -- & +1.94 & -4.74 & -8.50 & -8.77 & -2.75 & +2.36 & +9.79 & +2.12 & +4.65 & -0.18 \\
&
& \cellcolor{gray!18}\textsc{A}
& -0.08 & -0.03 & +0.00 & -0.12 & -0.12 & +0.22 & -0.03 & -0.48 & -3.32 & -2.01 & +0.49 & -6.15 & -17.81 & -15.37 & -3.69 & +1.77 & +4.07 & +0.70 & +1.92 & -0.45 \\
& \multirow{2}{*}{+DCO}
& \cellcolor{gray!10}\textsc{C}
& +0.20 & -- & +0.07 & +0.49 & +0.19 & +0.10 & +0.30 & +0.05 & +6.20 & -- & +4.99 & +6.66 & +5.28 & +6.56 & +7.03 & +5.33 & +13.68 & +4.63 & +6.45 & +1.35 \\
&
& \cellcolor{gray!18}\textsc{A}
& -0.20 & +0.03 & -0.23 & +0.10 & -0.32 & +0.04 & -0.15 & -0.88 & +0.88 & +0.53 & -0.57 & +1.73 & -0.06 & +0.66 & +1.84 & +0.21 & +2.95 & +1.35 & +1.45 & -0.45 \\
& \multirow{2}{*}{+INCLINE}
& \cellcolor{gray!10}\textsc{C}
& +0.26 & -- & +1.37 & -0.26 & +0.13 & -0.06 & +0.36 & -0.01 & -0.05 & -- & +0.04 & +0.14 & -0.24 & +0.19 & +0.42 & -0.28 & -0.63 & +0.08 & -0.46 & +0.22 \\
&
& \cellcolor{gray!18}\textsc{A}
& +0.02 & -0.02 & +0.15 & +0.15 & +0.17 & -0.12 & +0.08 & -0.28 & -0.16 & +0.07 & -0.07 & +0.06 & -0.09 & +0.10 & +0.04 & -0.04 & -0.92 & -0.19 & -0.94 & +0.19 \\
\cmidrule(lr){1-23}
\multirow{8}{*}{\texttt{Gemma3}}
& \multirow{2}{*}{Base}
& \cellcolor{gray!10}\textsc{C}
& 67.72 & -- & 69.55 & 73.20 & 65.15 & 65.20 & 68.08 & 65.11 & 67.39 & -- & 68.08 & 68.13 & 73.20 & 69.55 & 71.31 & 65.20 & 72.49 & 65.15 & 63.16 & 57.66 \\
&
& \cellcolor{gray!18}\textsc{A}
& 51.58 & 57.63 & 53.11 & 54.36 & 47.85 & 48.61 & 50.58 & 48.95 & 50.81 & 57.63 & 50.58 & 52.57 & 54.36 & 53.11 & 53.44 & 48.61 & 52.52 & 47.85 & 48.04 & 40.20 \\
& \multirow{2}{*}{+EN-A}
& \cellcolor{gray!10}\textsc{C}
& +0.21 & -- & -0.02 & +0.13 & +0.45 & +0.29 & +0.25 & +0.17 & +1.01 & -- & -0.84 & +2.16 & -0.36 & +2.15 & -0.71 & +0.38 & +0.82 & +0.63 & +4.01 & +1.89 \\
&
& \cellcolor{gray!18}\textsc{A}
& +0.06 & -0.03 & -0.20 & +0.00 & -0.01 & +0.13 & +0.31 & +0.22 & -0.01 & +0.13 & -0.14 & -1.35 & +0.14 & -1.04 & -0.25 & +0.24 & +0.98 & +0.66 & +0.34 & +0.20 \\
& \multirow{2}{*}{+DCO}
& \cellcolor{gray!10}\textsc{C}
& +0.75 & -- & +1.42 & +1.30 & +0.59 & +1.58 & -0.20 & -0.20 & +7.15 & -- & +6.00 & +7.65 & +6.97 & +7.96 & +6.67 & +6.63 & +6.61 & +6.23 & +8.50 & +8.32 \\
&
& \cellcolor{gray!18}\textsc{A}
& +0.02 & +0.06 & -0.01 & -0.10 & +0.16 & -0.10 & +0.35 & -0.22 & -2.27 & -1.98 & -2.34 & -1.05 & -3.68 & -2.00 & -0.96 & -2.37 & -1.58 & -3.17 & -2.47 & -3.42 \\
& \multirow{2}{*}{+INCLINE}
& \cellcolor{gray!10}\textsc{C}
& +0.34 & -- & +0.88 & +0.63 & +0.17 & -0.04 & +0.20 & +0.22 & -0.11 & -- & -0.16 & -0.17 & -0.19 & +0.26 & -0.02 & -0.15 & -0.35 & -0.31 & +0.09 & -0.07 \\
&
& \cellcolor{gray!18}\textsc{A}
& +0.08 & +0.08 & -0.03 & +0.08 & +0.01 & +0.24 & -0.05 & +0.22 & +0.03 & +0.07 & +0.17 & -0.01 & -0.08 & -0.17 & -0.10 & +0.27 & +0.28 & -0.08 & -0.02 & +0.04 \\
\cmidrule(lr){1-23}
\multirow{8}{*}{\texttt{Aya}}
& \multirow{2}{*}{Base}
& \cellcolor{gray!10}\textsc{C}
& 71.74 & -- & 76.02 & 77.36 & 69.59 & 69.07 & 69.88 & 68.52 & 69.51 & -- & 69.88 & 74.31 & 77.36 & 76.02 & 76.28 & 69.07 & 71.12 & 69.59 & 65.09 & 46.36 \\
&
& \cellcolor{gray!18}\textsc{A}
& 54.16 & 59.79 & 55.76 & 56.36 & 50.83 & 52.02 & 52.93 & 51.40 & 51.81 & 59.79 & 52.93 & 54.14 & 56.36 & 55.76 & 54.63 & 52.02 & 55.03 & 50.83 & 46.59 & 31.88 \\
& \multirow{2}{*}{+EN-A}
& \cellcolor{gray!10}\textsc{C}
& -0.06 & -- & -0.15 & -0.27 & -0.19 & +0.24 & -0.13 & +0.16 & -1.70 & -- & -1.02 & -1.81 & -0.09 & -0.12 & -0.61 & -3.13 & -1.77 & -1.73 & -3.26 & -3.48 \\
&
& \cellcolor{gray!18}\textsc{A}
& +0.08 & +0.12 & +0.09 & +0.03 & -0.13 & +0.16 & +0.09 & +0.18 & -1.26 & -0.16 & -0.79 & -1.24 & -1.65 & -0.54 & +0.09 & -1.74 & -1.19 & -1.17 & -1.41 & -4.08 \\
& \multirow{2}{*}{+DCO}
& \cellcolor{gray!10}\textsc{C}
& +0.13 & -- & +0.37 & +0.02 & +0.20 & +0.04 & +0.14 & +0.01 & +3.96 & -- & +3.90 & +3.91 & +2.48 & +2.92 & +2.99 & +3.52 & +7.85 & +2.20 & +2.79 & +6.99 \\
&
& \cellcolor{gray!18}\textsc{A}
& +0.06 & +0.15 & +0.11 & +0.07 & +0.06 & -0.03 & +0.07 & -0.04 & +0.51 & +0.94 & +0.44 & +0.45 & +0.37 & +0.52 & +0.82 & +0.66 & +0.39 & +0.01 & +0.64 & +0.35 \\
& \multirow{2}{*}{+INCLINE}
& \cellcolor{gray!10}\textsc{C}
& +26.50 & -- & +21.24 & +19.10 & +29.55 & +30.93 & +30.12 & +28.07 & -0.06 & -- & +0.01 & +0.03 & +0.03 & -0.08 & -0.05 & +0.18 & -0.27 & -0.16 & +0.12 & -0.39 \\
&
& \cellcolor{gray!18}\textsc{A}
& -27.19 & -33.11 & -28.71 & -26.97 & -26.38 & -24.97 & -25.87 & -24.35 & -0.00 & -0.00 & +0.19 & -0.01 & -0.02 & -0.03 & +0.06 & +0.13 & +0.06 & -0.06 & -0.21 & -0.15 \\
\bottomrule
\end{tabular}%
}
\caption{Cross-domain transfer results on MMMLU after post-training on BMLAMA or XCSQA. M. denotes the evaluation metric. \protect\colorbox{gray!10}{\textsc{C}} and \protect\colorbox{gray!18}{\textsc{A}} denote consistency and accuracy, respectively. Baseline rows report absolute scores, while method rows report absolute changes relative to the corresponding baseline. EN-A abbreviates EN-Align.}
\label{tab:mmmlu_transfer_bmlama_xcsqa}
\endgroup
\end{table*}

\subsection{Additional Linear-Probing Results}
\label{app:additional_probe_results}

\begin{figure*}[!h]
    \centering
    \includegraphics[width=0.9\linewidth]{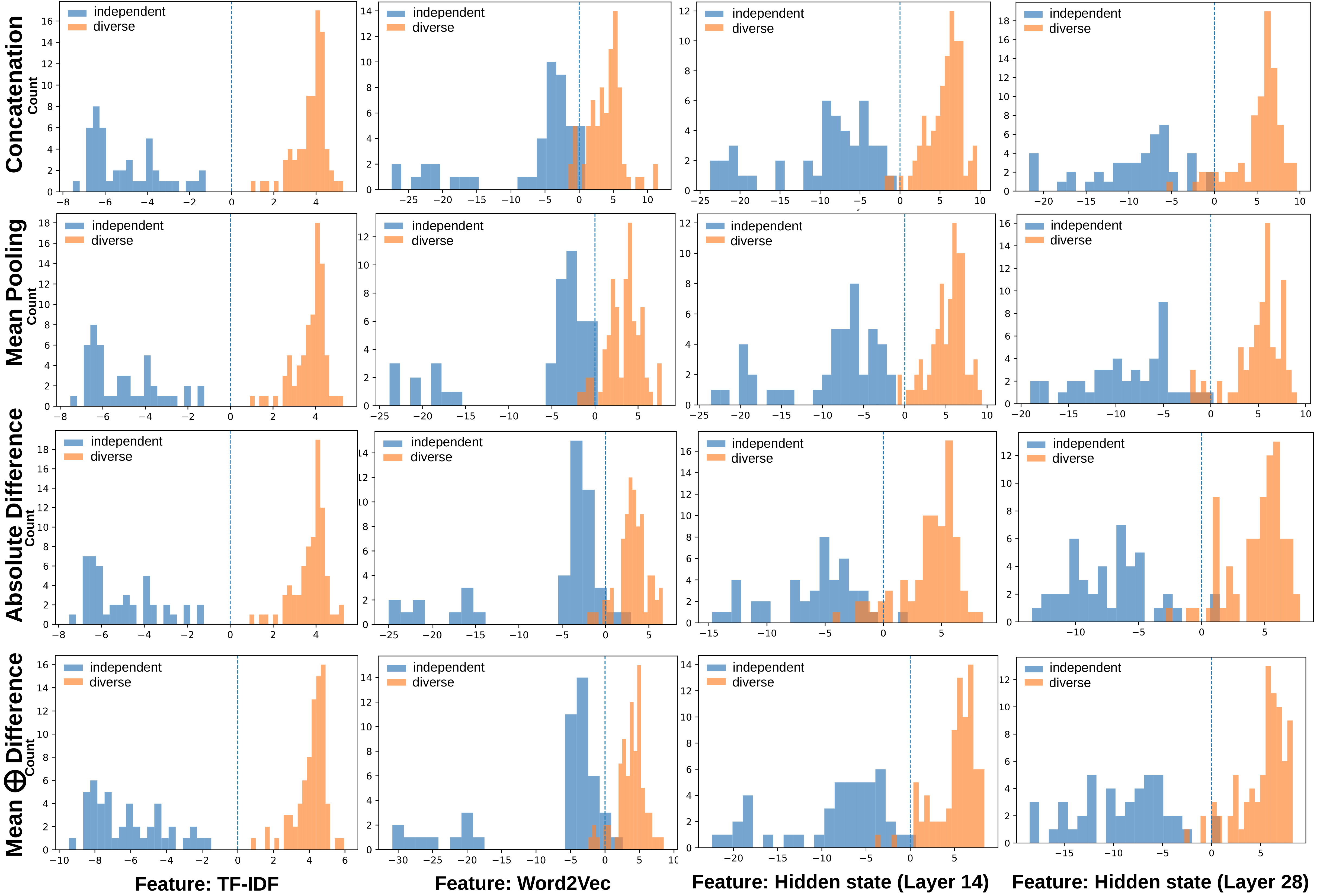}
    \caption{Linear-probe decision scores for four combination strategies on English–Hindi query pairs using Qwen2.5. Across TF-IDF, Word2Vec, and middle- and final-layer hidden-state features, all combination strategies show a consistent separability pattern between culture-independent and culture-diverse query pairs.}
    \label{fig:full_strategies}
\end{figure*}

\end{document}